\documentclass[10pt,twocolumn,letterpaper]{article}

\usepackage{cite}
\usepackage{amsmath,amssymb,amsfonts}
\usepackage{algorithmic}
\usepackage{textcomp}
\usepackage{xcolor}
\usepackage{multirow}
\usepackage{iccv}
\usepackage{times}
\usepackage{epsfig}
\usepackage{graphicx}
\usepackage{amsmath}
\usepackage{amssymb}

\usepackage{xcolor}

\usepackage[breaklinks=true,bookmarks=false]{hyperref}

\iccvfinalcopy 

\ificcvfinal\fi
\def\BibTeX{{\rm B\kern-.05em{\sc i\kern-.025em b}\kern-.08em
    T\kern-.1667em\lower.7ex\hbox{E}\kern-.125emX}}
\begin{document}

\title{Masked Feature Modelling: Feature Masking for the unsupervised pre-training of a Graph Attention Network block for bottom-up video event recognition
\thanks{This work was supported by the EU Horizon 2020 programme under grant agreement 101021866 (CRiTERIA).}}


\author{Dimitrios Daskalakis, Nikolaos Gkalelis\thanks{Work done while at CERTH-ITI.}, Vasileios Mezaris \\
CERTH-ITI \\
Thessaloniki, Greece, 57001 \\
\{dimidask, gkalelis, bmezaris\}@iti.gr }

\maketitle

\ificcvfinal\thispagestyle{empty}\fi

\begin{abstract}
In this paper, we introduce Masked Feature Modelling (MFM), a novel approach for the unsupervised pre-training of a Graph Attention Network (GAT) block. MFM utilizes a pretrained Visual Tokenizer to reconstruct masked features of objects within a video, leveraging the MiniKinetics dataset. We then incorporate the pre-trained GAT block into a state-of-the-art bottom-up supervised video-event recognition architecture, ViGAT, to improve the model's starting point and overall accuracy. Experimental evaluations on the YLI-MED dataset demonstrate the effectiveness of MFM in improving event recognition performance.
\end{abstract}

\section{Introduction}
\label{sec:intro}
\begin{figure*}[!ht]
\centering
\includegraphics[width=2.1\columnwidth]{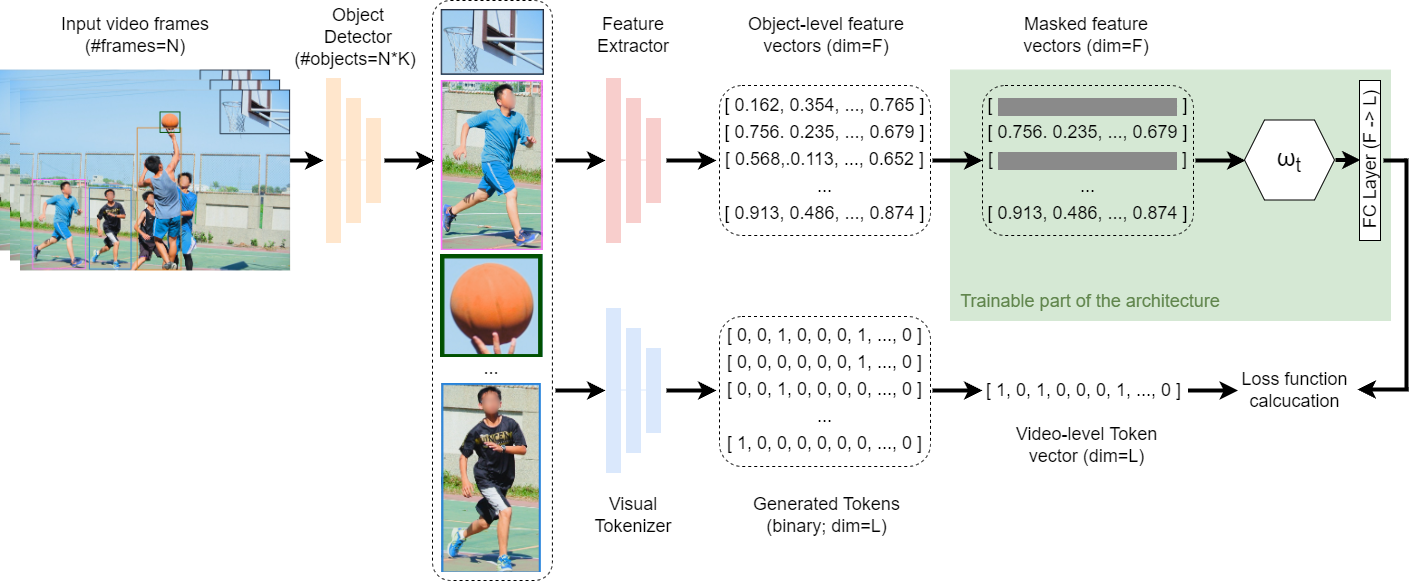}
\caption{An illustration of the main contribution of our paper, training a Graph Attention (GAT) block, $\omega_t$ \cite{Gkalelis2022} through Masked Feature Modelling and Tokens. Initially, we detect the objects within a video frame, pass them through the feature extractor to acquire object-level features. In parallel, a Tokenizer creates object-level token vectors of the objects based on their cosine similarity to a predefined Visual Vocabulary, to subsequently generate a video-level token vector, which is used as supervision in our unsupervised architecture. In each training iteration, a portion (e.g. 40\%) of the extracted object-level features are masked and the modified set of features is fed through $\omega_t$ and a fully connected (FC) layer. The binary cross-entropy loss is used for the unsupervised training, with the help of the generated token, of the trainable components of this architecture: $\omega_t$ and $FC$.}
\label{fig:unsupervised}
\end{figure*}

In recent years, Vision Transformers (ViTs) have emerged as a dominant approach in video and image analysis, gradually supplanting Convolutional Neural Networks in various applications. Such techniques have garnered significant scientific interest, with a constant influx of studies. However, one key challenge faced by ViTs is their requirement for abundant data and extensive annotations to achieve optimal training results.
To tackle this annotation-hungry nature of ViTs, several techniques were studied, such as Transfer Learning and, more recently, Masked Image Modelling (MIM).
The primary objective of MIM is to learn to reconstruct the masked patches of an image in order to capture comprehensive contextual information using a representation model. 
For example, in the case of BEiT \cite{Bao2022beit}, the pretraining process involves utilizing two different views of each image: image patches and visual tokens. 
Initially, the original image is converted into a set of discrete tokens, corresponding to an image patch, by encoding it into vectors and looking up its nearest neighbor in a "visual vocabulary". Then, randomly selected image patches are masked and both masked and unmasked image patches are fed into ViTs.
The objective of the pretraining procedure involves recovering the original visual tokens based on the masked image patches.
This pretraining does not require any ground-truth annotations for the employed image corpus.
Subsequently, the pretrained vision encoder can be deployed and fine-tuned for various downstream tasks by appending lightweight task-specific layers.
Although many studies in the image domain have experimented with this masking approach, masking has not yet been thoroughly explored in the realm of video event recognition or video-related tasks in general.
Given the inherent characteristics of this field, leveraging MIM approaches can facilitate the generalization of training models.
This, in turn, enables knowledge transfer across datasets, particularly from large-scale to smaller-scale datasets.
Moreover, MIM approaches can alleviate, to some extent, the challenges posed by the substantial computational and data requirements typically associated with achieving state-of-the-art results in video analysis. 

Self-supervised pre-training (including, but not limited to MIM) is in general a promising direction for addressing the annotation-hungry nature of ViTs. Such techniques enable models to learn from unlabeled data and capture underlying patterns and representations. 
For instance, Goulao \cite{goulao2022pretraining} introduced a temporal order verification task to enhance the capturing of temporal relations between observations, thereby improving the performance of Vision 
Transformers in Reinforcement Learning tasks. Furthermore, a more comprehensive study on self-supervised pre-training is presented in \cite{atito2021sit}. 
Despite the thorough exploration of unsupervised pre-training in various domains, a significant research gap remains in applying such techniques specifically to Graph Attention Networks that build upon feature embeddings, which have been shown to be key components of a state-of-the-art bottom-up video event recognition method \cite{Gkalelis2022}. Our work contributes to reducing this gap.

We present a new approach called Masked Feature Modelling (MFM), particularly tailored to videos (Fig.\ref{fig:unsupervised}).
In summary, our major contributions are: 

\begin{itemize}
\item We are the first, to the best of our knowledge, that apply vector-quantized visual tokenizer MFM techniques \cite{peng2022beit} to extracted object features within videos for unsupervised pretraining of Graph Attention Networks.
\item We show that the Graph Attention Networks pretrained through the proposed MFM technique can provide improvements in high-level event recognition accuracy in video.
\end{itemize}

\section{Related Work}
\label{sec:RelatedWork}
We focus the survey of the related work on the three domains that are most closely related to his work: a) Transfer Learning and Unsupervised Pre-Training, b) Masked Image Modelling and c) Video Event Recognition.

\subsection{Transfer Learning and Unsupervised Pre-Training}
\label{ss:TLVRrelated}
Transfer Learning (TL) refers to the process of taking advantage of the learned feature maps from a model trained on a large dataset and applying them to another neural network model designed for similar but relatively data-limited tasks. 
This approach offers several notable advantages, including reduced training time, improved performance, and less reliance on extensive labeled data.

When discussing TL, there are three primary approaches to consider. 
The first approach is Fixed Feature Extraction, which involves removing the classification head from a pretrained network while preserving the remainder of the network.
The second approach is Fine Tuning, which involves replacing the classification head of a pretrained model with a new one, suitable for the downstream classification task, and fine-tuning it using the downstream dataset to achieve improved accuracy. 
The third approach concerns weight initialization, wherein a model is initially trained on a larger dataset and subsequently retrained on a smaller one.
This methodology leverages the knowledge gained from the broader dataset to provide a solid foundation for further refinement and adaptation on the specific requirements of the smaller dataset.

The latter approach, although not extensively studied in various areas, has shown promise in different domains. 
Researchers have recently begun exploring the potential applications of unsupervised pre-training in various fields.
For instance, in tumor classification, an Autoencoder has been trained on a large dataset of tumor images, allowing the model to learn and capture meaningful representations \cite{MafaldaBIBM2018}. 
Additionally, Echo State Networks, a type of Recurrent Neural Networks, have shown advantages in tasks involving temporal dependencies and memory, such as time series analysis and prediction \cite{SteinerNNLS2022}.
Furthermore, weight initialization has been applied in Neural Network for Neuro PID Controllers \cite{TheerthamICT2018} and audio surveillance of roads employing deep learning techniques \cite{ZiedMELECON2020}. 
Notably, the latter study represents the sole research endeavor we came across that specifically focuses on weight initialization in video applications. This finding pinpoints the limited exploration of weight initialization in the realm of videos, underscoring the potential for further investigation and advancements in this particular domain.

Unsupervised pretraining is a powerful technique in deep learning that involves training models on unlabeled data before fine-tuning on labeled data for specific tasks. 
Regarding video classification, \cite{MilacskiIJCNN2019} utilizes unsupervised pretraining to learn temporal dependencies and extract discriminative features from videos. 
For person re-identification, an intra-identity regularization approach is proposed by  \cite{YangCVPR2022}. 
This technique employs unsupervised pretraining to enhance the model's ability to differentiate between individuals across multiple camera views. 
In the context of fully convolutional neural networks (FCNs), \cite{WiehmanPRASA2016} demonstrates the benefits of unsupervised pretraining for semantic segmentation tasks. 
By training FCNs on large amounts of unlabeled data, the models can learn meaningful features and improve their generalization capabilities before being fine-tuned on labeled data for specific segmentation tasks. 
\begin{figure*}[!ht]
\centering
\includegraphics[width=1.60\columnwidth]{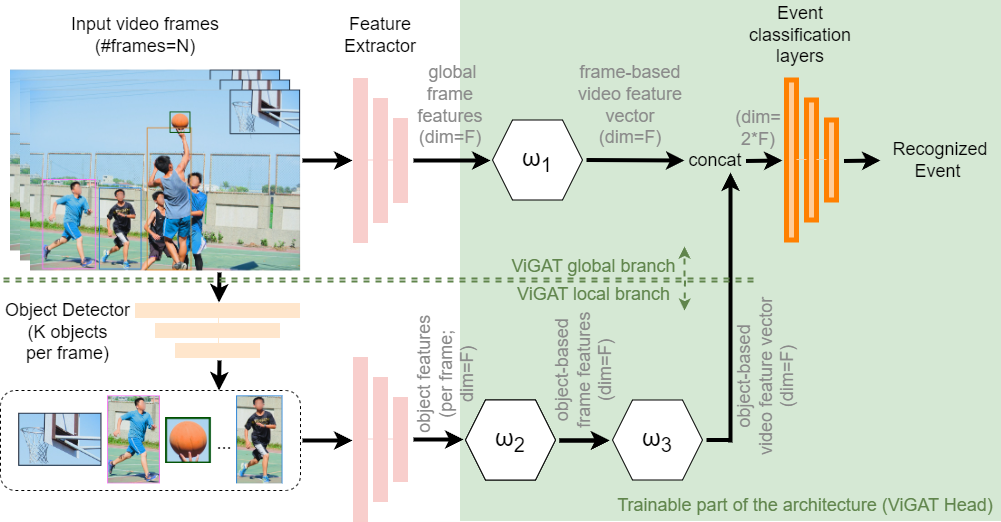}
\caption{The block diagram of ViGAT \cite{Gkalelis2022}. We adopt this method for the supervised video event recognition task. It encompasses an object detector, a feature extractor, and the ViGAT head. The ViGAT head consists of three Graph Attention Network (GAT) blocks ($\omega_1$, $\omega_2$, $\omega_3$) responsible for processing global (frame-level) and local (object-level) features. Finally, a pair of event classification layers utilize the video-level feature vectors coming out of GAT blocks $\omega_1$ and $\omega_3$ to recognize the event occurring in the video.}
\label{fig:supervised} 
\end{figure*}

\subsection{Masked Image Modelling}
\label{ss:MIMrelated}

Masked Image Modeling (MIM) is a technique that leverages the reconstruction of masked image content to learn new representations. Its early development can be traced back to approaches such as context encoder \cite{Pathak2016CVPR} and denoising autoencoder \cite{Fedus2022MLR}. 
More recently, this approach has been revisited and applied in training ViTs, as seen in BEiT \cite{Bao2022beit}, which introduces the prediction of discrete visual tokens as a key element, and iGPT \cite{Chen2020ICML} that adopts a sequential auto-regressive prediction of pixels. 

SimMIM  \cite{Xie2022CVPR} highlights the effectiveness of using raw pixel values from randomly masked patches as a reconstruction target during pretraining. It demonstrates that a lightweight classification head is sufficient for achieving satisfactory results.
Conversely, MAE \cite{He2022CVPR} adopts a different approach by considering only the visible patches as input to the encoder. 
In MAE, mask tokens are incorporated within the encoder and decoder, resulting in an asymmetric design that significantly reduces the computational overhead of the encoder. 
Furthermore, to enhance the feature extraction capability of the encoder, CAE \cite{Chen2022CAE} introduces a feature alignment module that explicitly separates the encoder and decoder. 
This explicit separation improves the encoder's ability to extract meaningful features. 
In the domain of hierarchical ViTs, there are concurrent efforts exploring the application of MAE. UM-MAE \cite{Li2022uniform} introduces a novel masking strategy tailored for pretraining pyramid-based ViTs.
Similarly, GreenMIM \cite{Huang2022green} adapts MAE for hierarchical architectures by dividing local windows into multiple equally-sized groups and proposing an optimal grouping algorithm to determine the ideal group size. 
MixMAE \cite{Liu2022mixmim} focuses on rearranging the inputs and targets, rather than designing specific masking or grouping strategies. 

Lastly, in the video domain, OmniMAE \cite{Girdhar2023CVPR} jointly trained a self-supervised ViT model with images and videos using sample replication and higher masking ratios.
In contrast, MaskFeat \cite{Wei2022CVPR} employs Histograms of Oriented Gradients to estimate the characteristics of the masked regions. 
On the other hand, MAGVIT \cite{Yu2023cvpr} introduces a Conditional Masked Modeling approach that replaces the conventional masking token, employing specific conditions as its foundation. 
It is evident that the studies conducted in the video domain are relatively limited compared to the extensive research carried out in the image domain. Moreover, a notable gap exists as no previous studies have actively explored the application of feature masking techniques in either images or videos. This observation served as a strong motivation for our research, as we tried to bridge this gap and investigate the synergistic potential of combining weight initialization and MFM specifically for video event recognition.

\subsection{Video Event Recognition}
\label{ss:EventRec}
In recent years, two dominant directions have been explored in this domain. The first direction involves training a Transformer or similar network from scratch using images or video frames. 
For instance, VTN \cite{neimark2021video} adopts a sliding window attention approach and a dedicated classification token at the beginning of the feature sequence, drawing inspiration from BERT \cite{devlin2018bert}. 
Another example is TimeS-Former \cite{bertasius2021space}, which utilizes frame-level patches to learn spatiotemporal features. It employs "divided attention" within each Transformer block, separating spatial and temporal attention, to classify videos.
Video Vision Transformer \cite{Arnab_2021_ICCV} employs a factorized attention mechanism to effectively handle spatial and temporal dimensions, enabling efficient processing of long video sequences.
However, these methods require training on very large annotated datasets and suffer from high computational complexity.

To reduce computational cost, a second direction involves utilizing pretrained models to extract feature representations, which are then fed into the respective classifier models and used for efficient video categorization.
AdaFocusV3 \cite{wang2022adafocusv3} utilizes a differentiable interpolation-based patch selection operation together with 3D video cubes to perform spatial-temporal dynamic computation simultaneously. In \cite{xia2022nsnet}, NSNet suppresses the responses of non-salient frames through the use of pseudo labels on the frame-level and dual video-level supervision on the video-level. In \cite{lvu2021}, an object-centric transformer is pretrained in self-supervised manner and fine-tuned for long-form video understanding. Lastly, ViGAT \cite{Gkalelis2022}, as depicted in Fig.\ref{fig:supervised}, and its faster approximation Gated-ViGAT \cite{gatedvigat22}, incorporate an object detector to extract bottom-up (object) information from video frames. The extracted objects and frames are then processed by a pretrained ViT backbone \cite{DosovitskiyICLR21} to derive feature representations. Finally, an attention-based head network which consists of three Graph Attention Network (GAT) blocks is employed to recognize and explain events within the video. These GAT blocks, namely $\omega_1$, $\omega_2$ and $\omega_3$, accept feature representations as input and use two graph layers to produce video feature vectors. With the use of two event classification layers, these video feature vectors provide the recognition result of a video. This holistic approach allows ViGAT and Gated-ViGAT to capture both object-level and frame-level information. 

In this paper, our goal is to (pre-)train the GAT blocks of ViGAT (Fig.\ref{fig:supervised}) without supervision. Building upon the weight-sharing technique described in \cite{Gkalelis2022}, we accomplish this by training a new GAT block $\omega_t$ using local (i.e. object) information and without supervision; and, subsequently, utilizing the pretrained $\omega_t$ to initialize one or more GAT blocks of ViGAT. We achieve this unsupervised training, on a large source-dataset, through the application of masked feature modelling and the utilization of the Vision Tokenizer introduced in \cite{Bao2022beit}. Once we have successfully completed this unsupervised training procedure, we initialize the selected GAT blocks of ViGAT with our $\omega_t$ and conduct supervised training and evaluation in a video-event recognition task on a small target-dataset. We should note that the object-centric transformer presented in \cite{lvu2021} also utilizes object features and a masked unsupervised pretraining step. However, in contrast to \cite{lvu2021} that utilizes a self-supervised contrastive loss (InfoNCE \cite{Oord2018}) at scene level, here we use a pretrained vector-quantized visual tokenizer with a cross-entropy loss \cite{peng2022beit}. The latter approach (which we use here to pretrain our Graph Attention Network) puts more emphasis on local- (in our case object-) level reconstruction. This design choice of our approach is further motivated by recent results in the image classification domain, which showed that BEiT v2 (utilizing a vector-quantized visual tokenizer) clearly outperforms MoCo v3 that is based on the InfoNCE loss \cite{ChenXH21} (see Table 2 in \cite{peng2022beit}).

\section{Proposed Method}
\label{sec:ProposedMethod}

Our method comprises an object detector, a feature extractor, a Visual Tokenizer, and a GAT (Graph Attention Network) block referred to as $\omega_t$. The GAT block $\omega_t$ adopts the structure defined in \cite{Gkalelis2022}, i.e. it is made of an attention mechanism, a GAT head of two layers, and a graph pooling stage; the interested reader is referred to \cite{Gkalelis2022} for more details on the GAT block structure. This $\omega_t$ is trained using tokens generated from a Visual Tokenizer \cite{peng2022beit} in conjunction with a novel approach based on MIM that utilizes features instead of images. This unsupervised procedure is called Masked Feature Modelling (MFM).

To acquire object-level features, we utilize the method employed in \cite{Gkalelis2022}. That is, a video is represented with a sequence of $N$ frames, and an object detector along with a feature extractor is used to obtain a matrix $\mathbf{X}^{(n)} \in \mathbb{R}^{K \times F}$ that represents frame $n$,
\begin{equation}
\mathbf{X}^{(n)} = [\mathbf{x}_1^{(n)}, \dots, \mathbf{x}_K^{(n)}]^T, \label{E:feaMatLocal}
\end{equation}
where $K$ is the number of objects extracted from each frame and $\mathbf{x}_k^{(n)} \in \mathbb{R}^F$ is the feature embedding for object $k$ in frame $n$.

In order to achieve unsupervised learning, we made the decision to utilize the pretrained Visual Tokenizer \cite{peng2022beit}.
As can be seen in Fig.\ref{fig:unsupervised}, by feeding an object image into the Tokenizer, we obtain new representations, called tokens, that serve as valuable supervision for our procedure.
The Tokenizer consists of a vision Transformer encoder, a quantizer and a visual vocabulary (Codebook) containing $L$ distinct embeddings.
Inside the Tokenizer, the object images are partitioned to $Q$ patches, and each patch is transformed to an embedding $\mathbf{h}_{j,k}^{(n)}$, which corresponds to the $j$th patch of the $k$th object of frame $n$. Then the quantizer looks up the nearest neighbor in the visual vocabulary for each representation $\mathbf{h}_{j,k}^{(n)}$, according to cosine similarity, and thereby produces the respective visual token vector $\mathbf{z}_{j,k}^{(n)} = [z_{1,j,k}^{(n)}, \dots, z_{L,j,k}^{(n)}]^T$, where, $z_{i,j,k}^{(n)} \in \{ 0, 1 \}$, and $z_{i,j,k}^{(n)}$ equals 1 if the $j$th patch of the $k$th object in the $n$th frame belongs to the $i$th codebook embedding, and 0 otherwise. A visual token vector $\mathbf{v}$ for the overall video is then obtained using
\begin{equation}
    \mathbf{u} = \sum_{n=1}^{N} \sum_{k=1}^{K} \sum_{j=1}^{Q} \mathbf{z}_{j,k}^{(n)},
\end{equation}
and the function $\mathsf{top}_r$ that returns 1 for the $r$ largest elements of $\mathbf{u}$ and 0 for the rest,
\begin{equation}
    \mathbf{v} = \mathsf{top}_r (\mathbf{u}).
\end{equation} 

To proceed with the unsupervised training, given the feature matrix $\mathbf{X}^{(n)}$, we mask $\Gamma \%$ of the objects in each frame
\begin{equation}
\mathbf{x}_k^{(n)} = \delta (k \in \mathcal{M}) \mathbf{p} + (1 - \delta (k \in \mathcal{M})) \mathbf{x}_k^{(n)},
\end{equation}
where, $\delta()$ is the indicator function, $\mathbf{p} \in \mathbb{R}^{F}$ is a shared learnable object feature embedding and $\mathcal{M}$ is the set of object indices $k$, randomly selected to be masked.
The resulting, so-called masked feature matrix $\tilde{\mathbf{X}}^{(n)}$, is processed by $\omega_t$ to produce in its output a latent representation; and then, a FC layer of $F$ inputs and $L$ outputs, equipped with an appropriate nonlinearity (i.e., softmax or sigmoid), is utilized to transform the output of $\omega_t$ to a score vector $\mathbf{g} \in \mathbb{R}^{L}$, containing $L$ score values with respect to the codebook vocabulary for the overall video.
The standard cross-entropy loss is then used to compute the  dissimilarity between $\mathbf{g}$ and the visual token vector $\mathbf{v}$.

In this way, using a large unlabelled video dataset, the GAT block $\omega_t$ can be trained effectively in an entirely unsupervised manner. Subsequently, it can be used for initializing and re-training a supervised event-recognition architecture (ViGAT) on a smaller target-dataset, to effect knowledge transfer. 

In the supervised event-recognition task, we utilize the pretrained $\omega_t$ to initialize one or both of $\omega_2$ and $\omega_3$ depicted in Fig.\ref{fig:supervised}. The entire ViGAT architecture is then trained on the YLI-MED dataset. Object and frame features are obtained using the object detector and feature extractor and are subsequently fed into the respective GAT blocks of the local and global branches. Finally, the video feature vectors are concatenated and passed through a classification head consisting of two fully connected layers to provide a prediction score for each event class.

\section{Experiments}
\label{sec:Experiments}
\subsection{Datasets}
\label{ssec:Datasets}
We conduct our experiments on two established and publicly available video datasets:

i) MiniKinetics \cite{XieECCV18} is a subset of the Kinetics dataset \cite{Carreira_2017_CVPR}.
It comprises 200 action classes, 121215 training and 9867 testing video clips. 
Each clip, sampled from a distinct YouTube video, has a duration of 10 seconds and is annotated with a single event/action class label. We utilize this as the source dataset, i.e. for training $\omega_t$ in an unsupervised way without using the existing class-label annotations, as illustrated in Fig.\ref{fig:unsupervised}.

ii) YLI-MED \cite{bernd2015yli} is a TRECVID-style video corpus based on YFCC100M, containing 1823 videos and 10 event categories. 
The dataset is divided into standard training and testing partitions of 1000 and 823 videos, respectively. We employ this much smaller dataset as the target one, i.e. for supervised learning, taking advantage of the previously trained $\omega_t$ as illustrated in Fig.\ref{fig:supervised}.

\subsection{Setup}
\label{ssec:Setup}

In order to accurately represent each video within the two datasets, we initially employ uniform sampling, resulting in a sequence of $N = 9$ or $N = 25$  frames (depending on the experiment) for YLI-MED and $N=30$ frames for MiniKinetics.
Our approach consists of the following:

a) An object detector named Detic \cite{zhou2022dedetic}, which is pretrained on ImageNet21K and fine-tuned on the CoCo dataset.

b) A ViT-L/14-Clip backbone, utilizing the OpenAI CLIP model \cite{radford2021learning} for the extraction of our object-level and frame-level features.
This ViT backbone utilizes a $14 \times 14$ partitioning grid to patchify the input image, i.e., $Q=14^2$ patches are produced for each input frame. Moreover, the dimension of the derived feature embeddings is $F=1024$.

c) The pretrained Visual Tokenizer provided in \cite{peng2022beit} with Codebook size $L = 8192$.

We should note that the original implementation of ViGAT in \cite{Gkalelis2022} utilized an older Faster R-CNN \cite{renNips2015faster} as the object detection component, and a lower-performing ViT-B/16 backbone for feature extraction.

For the object detection process, we set the number of objects $K$ to be extracted as 50. For the masking procedure we selected $\Gamma$ to be 40\%.
Our unsupervised architecture for pretraining $\omega_t$, depicted in Fig.\ref{fig:unsupervised}, underwent training on MiniKinetics for a total of 200 epochs. 
To facilitate convergence, we initially set the learning rate to $10^{-3}$, which was subsequently multiplied by 0.1 at epochs 50 and 100.

Our adopted supervised event recognition architecture, depicted in Fig.\ref{fig:supervised}, was trained on the YLI-MED dataset for 200 epochs.
The same training was performed on either the local branch alone or the entire architecture, depending on the experiment.
We adopted an initial learning rate of $10^{-4}$ and employed a learning rate schedule that involved multiplying it by 0.1 at epochs 60 and 110.

In alignment with established practices in the literature, we evaluate the recognition performance on the YLI-MED dataset utilizing the top-1 accuracy metric.

\subsection{Event recognition results}
\label{ssec:EventRecognitResults}

We evaluate the performance of our unsupervised pretrained GAT block $\omega_t$, by using it for the initialization of GAT blocks in our supervised event-recognition architecture ViGAT, on the YLI-MED dataset.
The upper part of Table \ref{tbl:ExpResTotal} shows our experiments using just the local branch of the entire ViGAT architecture, as depicted in Fig.\ref{fig:supervised}. 
To study the behavior of $\omega_t$ and minimize the influence of frame-level features, we deliberately selected a smaller number of frames in this experiment, specifically $N=9$. 
The results demonstrate that our pretrained $\omega_t$ outperforms the randomly initialized GAT blocks, achieving an improvement of 0.58\%. 

\begin{table}[t]
\centering
\begin{tabular}{llllc}   $N$&
       $\omega_1$ & $\omega_2$ & $\omega_3$& top-1(\%) \\

\hline
\parbox[t]{1mm}{\multirow{2}{*}{\rotatebox[origin=l]{0}{$9$}}}
    & - & Rand Init&Rand Init & 87.12 \\
    & - & Pretr. $\omega_t$&Pretr. $\omega_t$  & \textbf{88.70} \\
\hline
\parbox[t]{1mm}{\multirow{2}{*}{\rotatebox[origin=l]{0}{$25$}}}
    &Rand Init&Rand Init&Rand Init & 90.77 \\
    & Rand Init&Pretr. $\omega_t$&Pretr. $\omega_t$&\textbf{91.62} \\
\hline
\end{tabular}

\caption{Evaluation of the use of the unsupervised pretrained GAT block $\omega_t$ in our supervised event-recognition architecture, ViGAT (Fig.\ref{fig:supervised}), on YLI-MED, excluding or including ViGAT's global branch. The experiments are conducted using $N$ frames for each video.}
\label{tbl:ExpResTotal}
\end{table}
We also conducted another experiment where we used the complete ViGAT supervised event-recognition architecture of Fig \ref{fig:supervised}. In this experiment, we employed a total of $N = 25$ frames for both the local and global branches. The decision to increase the number of frames was motivated by the need to provide meaningful context and allow also the global branch to effectively learn. As presented in the lower part of Table \ref{tbl:ExpResTotal}, by utilizing the unsupervised pretrained $\omega_t$ for the initialization of the local-branch GAT blocks, ViGAT yields a significant improvement of 0.85\% compared to using randomly initialized GAT blocks. This highlights the effectiveness and potential of unsupervised pretraining in capturing meaningful representations, and enhancing the contribution of ViGAT's local branch to event recognition.

\subsection{Ablation Study}
\begin{table}[t]
\centering
\begin{tabular}{lllc}  
          &  & weight & \\
        $\omega_2$ & $\omega_3$& sharing & top-1(\%) \\
        
\hline
    Mean Pooling& Mean Pooling & no & 80.92 \\
    Pretrained $\omega_t$& Rand Init & no & 85.18 \\
    Rand Init&Mean Pooling & no & 86.51\\
    Rand Init&Rand Init & yes & 87.12 \\
    Pretrained $\omega_t$& Mean Pooling & no & 88.34\\
    Pretrained $\omega_t$& Pretrained $\omega_t$ & yes &\textbf{88.70} \\
\hline
\end{tabular}

\caption{Ablation study depicting the top-1 accuracy of the local branch of ViGAT on YLI-MED throughout different scenarios of using no trainable components, a pretrained $\omega_t$ or a randomly initialized GAT block.}
\label{tbl:AblationStudy}
\end{table}

In our ablation study, we compare different variants of our model by substituting the GAT blocks initialized with $\omega_t$ with alternative training approaches or a simple non-trainable mean pooling.
These comparisons specifically focus on the local branch of the ViGAT architecture of Fig.\ref{fig:supervised} (setting $N = 9$), in order to evaluate the effectiveness of knowledge transfer.
The results of these analyses are presented in Table \ref{tbl:AblationStudy}, where we observe:

a) Incorporating $\omega_t$ for initialization of $\omega_2$ and subsequent training on YLI-MED yielded a substantial improvement of 7.42\% compared to utilizing mean pooling alone for $\omega_2$, in the experiment where the trainable $\omega_3$ was substituted with mean pooling.

b) Random initialization of $\omega_2$ and training from scratch on YLI-MED also performs worse than using the pretrained $\omega_t$ for its initialization, when we retain mean pooling for $\omega_3$. In this case, the use of $\omega_t$ led to accuracy increase by approximately 2\%.

c) Utilizing $\omega_t$ for initializing both the objects- and the frames-GAT blocks, $\omega_2$ and $\omega_3$, where these two GAT blocks also share weights during the subsequent training on YLI-MED, is the best-performing strategy. It yields significantly improved results compared to either using a single randomly initialized GAT block or relying on randomly initialized GAT blocks for both objects and frames. This experiment shows improvements of 3.52\% and 1.58\%, respectively. 

These results highlight the advantages of introducing the proposed MFM approach for the unsupervised training of $\omega_t$, for improving the event recognition results of ViGAT on a small target dataset.

\section{Conclusion}
\label{sec:Conclusions}

In this paper, we introduced Masked Feature Modelling (MFM), and demonstrated the use of MFM for the unsupervised pre-training of a key component of the state-of-the-art ViGAT event recognition method. Our results showed that using the outcome of unsupervised pre-training for the initialization of certain blocks of ViGAT enables the latter to reach higher accuracy in a downstream task, i.e. when further trained in a supervised way on a small target dataset for event recognition. We believe that this work contributes to advancing our understanding of the benefits and applications of masking techniques in video analysis; and, that such Masked Feature Modelling can be widely applicable in various video classification problems, when the employed learning architecture leverages ``features'' (typically, visual information embeddings generated by pre-trained deep networks).

{\small

}

\end{document}